\documentclass[journal]{IEEEtran}

\usepackage{amsmath}
\usepackage{amssymb}
\usepackage{graphicx}
\usepackage{multirow}
\usepackage{amsfonts}
\usepackage{dsfont}
\usepackage{mathrsfs,amssymb}
\usepackage[english]{babel}
\usepackage[T1]{fontenc}
\usepackage{amsthm}
\usepackage{bbding}
\usepackage{epsfig}

\usepackage{color}

\usepackage{algorithm}
\usepackage{algorithmic}

\usepackage[font=small]{caption}
\usepackage{multicol}
\usepackage{balance}

 \usepackage{footmisc}

\begin{document}
\begingroup

\title{Cascaded Compressed Sensing Networks: A Reversible  Architecture for Layerwise Learning}

\author{Weizhi Lu, Mingrui Chen, Kai Guo and Weiyu Li

\thanks{W. Lu, M. Chen and K. Guo are with the School of Control Science and Engineering, Shandong University, Jinan, China. E-mail:  wzlu@sdu.edu.cn}
\thanks{W. Li is with the  Zhongtai Securities Institute for Financial Studies, Shandong University, Jinan, China. E-mail: liweiyu@sdu.edu.cn}

}

\maketitle




\begin{abstract}

Recently, the method that learns networks layer by layer has attracted increasing interest for its ease of analysis.   For the method, the main challenge lies in deriving an optimization target for each layer by inversely propagating the global target of the network. The propagation problem is   ill posed, due to involving the inversion of  nonlinear activations from low-dimensional to high-dimensional spaces. To address the problem, the existing solution  is to learn an auxiliary network to specially propagate the target. However, the network lacks   stability, and moreover, it  results in higher complexity for network learning. In the letter, we show that    target propagation could be    achieved  by  modeling the network's each layer with  compressed sensing, without the need of  auxiliary  networks. Experiments show that the proposed method could achieve better performance than the auxiliary network-based method.

\end{abstract}


\IEEEpeerreviewmaketitle

\theoremstyle{definition} \newtheorem{theorem}{Theorem}[]
\theoremstyle{definition} \newtheorem{lemma}[theorem]{Lemma}
\theoremstyle{definition} \newtheorem{definition}[theorem]{Definition}
\theoremstyle{definition} \newtheorem{property}[theorem]{Property}
\theoremstyle{definition} \newtheorem{corollary}[theorem]{Corollary}

\renewcommand{\thefootnote}{\fnsymbol{footnote}}

\section{Introduction}




The error backpropagation (BP) method has achieved great success in the supervised learning of deep neural networks \cite{AlexNet2012}. Due to taking the entire network as a whole to optimize, the method can hardly disclose the  contribution of each layer, causing difficulty to network analysis. As an alternative to BP, the emerging layerwise learning method \cite{bengio2014auto,lee2015difference} seems more suitable for network analysis. The method aims to learn the network layer by layer  and thus  reduces the network analysis  to the layer level.

For layerwise learning, the key is to derive an optimization target  for each layer, simply called  local target hereafter. Currently, the most simple method is to directly employ the global target \cite{marquez2018deep}, namely the optimization target of the entire network, as the  target of each layer.  However, the global target  cannot  guarantee to be optimal for each layer to minimize the global loss. To achieve BP level  performance, the methods in \cite{belilovsky2019greedy,nokland2019training}  propose to append a classifier to the layer of interest, and then optimize them together with the global target. This forms   a shallow network with at least one hidden layer, in which  the layer of interest is still hidden and thus hard to analyze \cite{barron1994approximation,pinkus1999approximation,bach2017breaking}.   To directly optimize and analyze each layer, we resort to another emerging layerwise learning method known as target propagation (TP) \cite{bengio2014auto,lee2015difference,bartunov2018assessing}, which derives the local target of each layer  by inversely propagating the global target, and  optimizes the layer by matching its output  to the  target. In the method,  local targets are required to be of two properties. First, their feedforward outputs  should be identical or close to the global target. This ensures that the optimization of each layer could reduce the global loss to the maximum extent. Second, the local target of each layer should be close to the forward output of the layer. This will help to reduce the variation of each layer over optimization, while the low variations incline to improve the network's generalization \cite{dziugaite2017computing,nagarajan2019generalization, trinh2019greedy}. However, the two properties are hard to achieve because the inverse propagation of global target  involves the inversions of nonlinear activations from low-dimensional to high-dimensional spaces. This is an  ill-posed problem. For the problem, the existing solution is to learn an auxiliary network to specially propagate the target. Obviously, the method will increase the  complexity of network learning, and moreover, the auxiliary network has no guarantee for stable and accurate target propagation.  In the letter, we show that auxiliary networks are not necessary for layerwise learning,  and  target propagation could be achieved with guaranteed accuracy by modeling the network's each layer with compressed sensing.

In compressed sensing \cite{Gribonval03,Donoho2003optimally},  it has been proved that a high-dimensional signal $x\in \mathbb{R}^n$ with sparse transformations $s=Dx$ over an orthonormal dictionary $D\in \mathbb{R}^{n\times n}$,   could be recovered from its low-dimensional projection $y=WDx=Ws$, where $W\in \mathbb{R}^{m\times n}$, $m\ll n$.  The recovery is implemented  by first recovering the sparse vector $s$ from $y$  via solving  an $\ell_0$  or $\ell_1$-regularized least square problem \cite{Chen01}, and then simply deriving $x=D^{\top}s$. Inspired by this reverse process, we propose to model each network layer to be a  compressed sensing process. Specifically, we formulate each layer as a cascade of $W$ and $D$, such that the input $s$ of $W$ could be recovered from its output $y$  via sparse recovery, and the input $x$ of $D$ (namely the output $y$ of $W$) could be simply recovered from its output $s$  by multiplying with $D^\top$. Repeating this process layer by layer, target propagation will be finally accomplished.  To obtain better performance, as usual,  we  suggest to further  deploy a  nonlinear function  $f(\cdot)$ following each dictionary  $D$, in order  to remove  small activations in magnitude. This operation brings two benefits. First, it further improves the sparseness of the output activations of $D$, which is beneficial to the sparse recovery on the following $\mathbf{W}$. Second, it tends to reduce  within-class scattering and thus improve the classification  accuracy \cite{Zarka2020Deep}.   Overall, with the perfect combination of $W$, $D$ and $f(\cdot)$, we establish a compressed sensing framework for each layer, and then obtain a cascaded compressed sensing architecture with all layers. To the best of our knowledge, this is the first time that compressed sensing is introduced for layerwise learning to handle the target propagation problem.

Compared to auxiliary networks-based TP methods, the proposed compressed sensing-based TP method has two major advantages. First, it avoids the usage of auxiliary networks and thus reduces the complexity of network learning. Second, it could achieve target propagation with guaranteed accuracy. In layerwise learning, the amount of  local targets we need to compute and store is  equal to  the number of  activations in the layer. Considering convolution networks contain huge amounts of activations and pose great computational challenges to layerwise learning,     as in  \cite{lee2015difference,bartunov2018assessing},  we only test the proposed  layerwise learning method  on fully-connected networks for tractable computation. Our performance advantage is verified by the classification experiments on MNIST and CIFAR10.

\section{Method}

  \begin{figure}[t]
\centering
 \graphicspath{fig}
\includegraphics[width=0.48\textwidth]{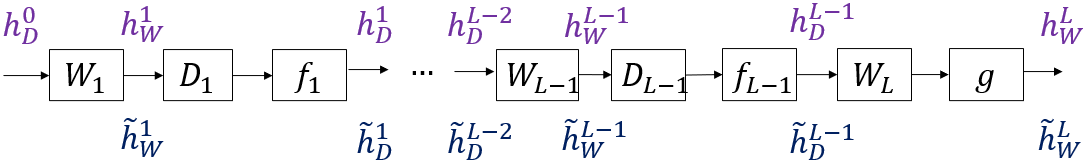}
\caption{The proposed depth-$L$ network structure, with feedforward activations  marked on the upper part and feedback targets marked   on the lower part.  \label{fig_network}}
\end{figure}

 Let us consider a deep fully connected network with $L$ layers as shown in Fig. \ref{fig_network}.   Each layer consists of two sublayers $\{W_l, D_l\}$, where  $W_l\in \mathbb{R}^{m_l\times n_l}$ with $m_l\ll n_l$ denotes a projection matrix and $D_l\in \mathbb{R}^{m_l\times m_l}$ means an orthonormal dictionary, $D^\top D=I$. Following $D_l$ is a pointwise thresholding function  $f_l(\cdot)$  which zeroes out all but the $k_l$ ($\ll m_l$) largest positive elements, acting similarly to the ReLU function. Note that the output layer is a classifier which comprises  $W_L$ but no $D_L$.   Suppose  $h_W^l\in \mathbb{R}^{m_l}$ and $h_D^l\in \mathbb{R}^{m_l}$ are the output activations of $W_l$ and $D_l$, respectively. The feedforward network structure is formulated by
 \begin{equation} \label{eq_feedforward}
 h_W^l=W_lh_D^{l-1}~~ \text{and}  ~~h_D^l=f_l(D_lh_W^l), ~~l=1,\ldots,L-1
 \end{equation}
 \begin{equation} \label{eq_feedforward2}
 h_W^l=g(W_lh_D^{l-1}), ~~l=L
 \end{equation}
where  $g(\cdot)$ is a softmax  function, and the notations  $h_D^{0}$ and $h_W^{L}$ denote the network's input and output, respectively.


As sketched in Algorithm \ref{alg-layerTrain}, the proposed layerwise learning is implemented within three phases. First, we initialize the network via feeding forward the network input $h_D^{0}$. During this process, the projection matrices $W_l$ are randomly generated, and the orthonormal dictionaries $D_l$ are learned with their input activations ${h}_W^{l-1}$. Second, as shown in the lower part of Fig.\ref{fig_network}, by compressed sensing the global targets $\tilde{h}_W^L$ are inversely propagated layer by layer,  producing local targets  $\tilde{h}_W^l$ and  $\tilde{h}_D^l$ respectively for $W_l$ and  $D_l$, $1\leq l<L$. Finally, we feed forward the input $h_D^{0}$ again, in order to successively minimize the local losses  $\mathcal{L}_l({h}_W^l$, $\tilde{h}_W^l)$ and  $\mathcal{L}_l({h}_D^l, \tilde{h}_D^l)$ of each layer by updating $W_l$ and $D_l$, $1\leq l<L$. This finally minimizes the global loss $\mathcal{L}({h}_W^L$, $\tilde{h}_W^L)$.  In the letter, we simply define the local losses  with Euclidean distances and the global loss with cross entropy. In the following, we detail the three phases.



    \begin{algorithm}[t]
\caption{ \label{alg-layerTrain} The proposed layerwise learning}
\begin{algorithmic}[1]
\STATE \emph{Phase 1: Activation forward propagation.}
\FOR {$l=1$ to $L$}
\STATE Generate $h_W^l$ and $h_D^l$ by \eqref{eq_feedforward} and \eqref{eq_feedforward2}.
\STATE Initialize $W_l$ with Gaussian distributions.
\STATE Learn $D_l$ with its input $h_W^l$ by Algorithm \ref{alg-DicLearn}.
\ENDFOR
\STATE Further update the final $W_L$ by \eqref{eq_InitialWL}.
\STATE \emph{Phase 2: Target backward propagation.}
\STATE Generate $\tilde{h}_D^{L-1}$ by \eqref{eq_hDp} and $\tilde{h}_W^{L-1}$ by \eqref{eq_backW}.
\FOR {$l=L-2$ to $2$}
\STATE Generate $\tilde{h}_D^{l-1}$ by \eqref{eq_backD} and $\tilde{h}_W^{l-1}$ by \eqref{eq_backW}.
\ENDFOR
\STATE  \emph{Phase 3: Forward layerwise  learning.}
\FOR {$l=1$ to $L-1$}
\STATE Update $W_l$ by \eqref{eq_updW} and $D_l$ by \eqref{eq_updD1} and  \eqref{eq_updD2}.
\ENDFOR
\STATE  Update the final $W_L$ by \eqref{eq_hWpd}.
\end{algorithmic}
\end{algorithm}

\subsection{Activation forward propagation} By  \eqref{eq_feedforward} and \eqref{eq_feedforward2}, we know the network takes $h_D^{0}$ as input and $h_W^{L}$ as output.  During the feedforward propagation, we initialize the projection matrix $W_l$  by drawing its elements from standard normal distribution. The distribution has been proved nearly optimal  for compressed sensing \cite{Candes05}. To keep the input activation $h_D^{l-1}$ unchanged  in $\ell_2$ norm  after projections, $W_l$ needs to be scaled by a factor $1/\sqrt{m_l}$ \cite{Jonson84}. For the output layer $W_L$, besides random initialization, we further update it   by
\begin{equation} \label{eq_InitialWL}
{W}_{L}={W}_{L}-\alpha\frac{\partial \mathcal{L}({h}_W^L, \tilde{h}_W^L)}{\partial {W}_{L}},
\end{equation}
in order to reduce the difference  between the final output $h_W^{L}$ and the global target $\tilde{h}_W^L$.  As detailed later, the reduced difference is beneficial to the final layerwise optimization. Empirically, the parameter $\alpha$ in \eqref{eq_InitialWL} should be small in case of overfitting.

 Next, we move to learning  the orthonormal dictionary $D_l$, which has  dense input  $h_W^l$ and sparse output $h_D^l$. Such kind of dictionaries can be learned  with an  procrustean approach \cite{schonemann1966generalized}, which is  introduced in Algorithm \ref{alg-DicLearn} for completeness. In the algorithm, there is a  crucial  parameter $d$, which counts the number of largest elements maintained by the thresholding function $T(\cdot)$. To obtain sparse transforms, we should adopt a relatively small $d$.  Note that  Algorithm \ref{alg-DicLearn} supports online learning, and the property is necessary for batch learning. From the viewpoint of classification, we can say that the sparse transform matrix $D_l$ is mainly used for feature selection, and the projection matrix $W_l$ serves not only for feature selection but for dimensionality reduction.  Moreover, it is noteworthy that the order of $D_l$ and $W_l$ could be reversed in the cascaded compressed sensing framework. In the letter, we decide to put  $W_l$ ahead in order to decrease the dimension of the network input  $h_D^0$ and then  reduce the optimization complexity.

Finally, let us see the initialization of the nonlinear function $f_l(\cdot)$, which has one thresholding parameter to maintain the $k_l$ largest positive elements in the output of $D_l$. Note that  $f_l(\cdot)$  only keeps a portion of positive elements and discards the negative ones, because empirically both of them tend to share the similar information for classification.   The choice of  $k_l$  depends on both the sparse degree of the output of $D_l$ and the compression ratio of the following $W_{l+1}$.   Empirically, as discussed later,  a large $k_l$ that allows for containing most of positive elements tends to provide good classification performance, when  the dictionary output is not very sparse.






\subsection{Target backward propagation}  The global target  $\tilde{h}_W^L$ is propagated backward by repeating the following compressed sensing recovery process:
\begin{equation}\label{eq_backD}
\tilde{h}_D^{l-1}=\arg\min_{{h}_D^{l-1}} \|\tilde{h}_W^l-W_l{h}_D^{l-1}\|_2^2 ~ \text{s.t.}~\|{h}_D^{l-1}\|_0\leq k_{l-1}
\end{equation}
and
\begin{equation}\label{eq_backW}
  \tilde{h}_W^{l-1}=D_{l-1}^\top \tilde{h}_D^{l-1}
\end{equation}
where the layer index $l$ decreases from $L$ to 2; $\tilde{h}_D^{l-1}$  and $\tilde{h}_W^{l-1}$ denote the  local targets derived respectively for  $D_{L-1}$ and $W_{L-1}$. The derivations of the two targets are analyzed as follows. Let us first see the deriving of  $\tilde{h}_D^{l-1}$  by $\eqref{eq_backD}$.  By compressed sensing, the row size $m_l$ of $W_l$ in \eqref{eq_backD} should be at least twice its input  sparsity $k_{l-1}$, namely $m_l>2k_{l-1}$,  in order to  approximately recover $\tilde{h}_D^{l-1}$ from  $\tilde{h}_W^{l}$. Empirically,  the condition requires  not to be strictly met, and a good classification is usually obtained as $m_l<2k_{l-1}$.  This implies that for layerwise learning, the feedforward of sufficient amounts of features may be more important than the accuracy of  target propagation.  We could solve  $\eqref{eq_backD}$ with many algorithms \cite{zhang2015survey}.  In our experiments, we  will employ the known  OMP  \cite{Pati93} algorithm, which has a parallel implementation in the software SPAMS \cite{mairal2014sparse}.  As for  the local target $\tilde{h}_W^{l-1}$, it is seen in \eqref{eq_backW} that the target could  be simply derived from $\tilde{h}_D^{l-1}$ by multiplication, benefiting from the orthonormality of $D_{l-1}$.


Usually, we prefer to simply label each class with a  one-hot vector $\tilde{h}_W^L$, such that the instances of the same class will share the same local target at each layer. Note that the label sharing may cause difficulty for layerwise optimization, as the instances of the same class have high  variations, as often encountered at early layers \cite{lee2015difference}. To avoid the problem, an effective solution is to adopt  diverse labels/tagets to represent the instances of the same class \cite{lee2015difference}. To achieve this,  instead of \eqref{eq_backD}, we propose to derive the local target $\tilde{h}_D^{L-1}$ of the penultimate layer by
\begin{equation} \label{eq_hDp}
\tilde{h}_D^{L-1}={h}_D^{L-1}-\beta\frac{\partial \mathcal{L}({h}_W^L, \tilde{h}_W^L)}{\partial {h}_D^{L-1}}
\end{equation}
 and then derive the   targets for the rest layers by \eqref{eq_backD} and \eqref{eq_backW}.


    \begin{algorithm}[t]
\caption{ \label{alg-DicLearn} Online orthonormal dictionary learning \cite{bao2013fast}}
\begin{algorithmic}[1]
\STATE Input: Dataset $X\in \mathbb{R}^{m\times n}$ of $n$ samples of length $m$ and  initialized dictionary $D^{(0)}$.
\FOR {$j=1$ to $J$}
\STATE $S^{(j)}=T({D^{(j)}}^{\top} X)$, $T(\cdot)$ is a  threshold function to zero out all but $d$ largest   elements (in magnitudes) in each column of its matrix input.
\STATE Run SVD for $X{S^{(j)}}^{\top}=P\Sigma Q^{\top}$.
\STATE $D^{(j+1)}=PQ^{\top}$.
\ENDFOR
\STATE Output: Learned dictionary $D={D^{(J)}}^{\top}$
\end{algorithmic}
\end{algorithm}

Considering the global loss $\mathcal{L}({h}_W^L, \tilde{h}_W^L)$ has been  previously  decreased in the initialization phase by updating $W_L$ in  \eqref{eq_InitialWL},  we use a small  $\beta$  in \eqref{eq_hDp} to further reduce the global loss to  achieve a good classification performance. It is seen that a small $\beta$ will result in a small difference ${h}_D^{L-1}-\tilde{h}_D^{L-1}$, and then a sequence  of small  ${h}_D^{l}-\tilde{h}_D^{l}$ (with $l$ decreasing from $L-2$ to $1$) by the stable recovery of compressed sensing  \cite{Candes06st}.  The small  differences between forward  activations and  backward targets will reduce the variations of  $D_{l}$ and $W_{l}$ in optimization, while the low variations over optimization incline  to improve the  network's generalization  \cite{dziugaite2017computing,nagarajan2019generalization, trinh2019greedy}. To achieve such property, the  TP method in \cite{lee2015difference}  constrains the difference between targets and activations, as  learning the auxiliary network. However, the  network-based method lacks stability, thus inferior to  our compressed sensing-based  method.




\subsection{Forward layerwise learning}
Feeding forward the  network input $h_D^0$, we successively update the parameters $\{{W}_l, {D}_l\}$  of each layer    by matching their outputs ${h}_W^{l}$ and ${h}_D^{l}$  to their local  targets $\tilde{h}_W^{l}$ and $\tilde{h}_D^{l}$.  Formally, this process is realized by iterating
\begin{equation}\label{eq_updW}
\tilde{W}_{l}=\arg\min_{W_{l}} \|\tilde{h}_W^l-W_l{h}_D^{l-1}\|_2^2+\lambda_{W_l}\|W_l\|_2^2
\end{equation}
\begin{equation}\label{eq_updWh}
{h}_W^{l}=\tilde{W}_{l}{h}_D^{l-1}
\end{equation}

 \begin{equation}\label{eq_updD1}
 {h}_W^{l} \tilde{h}{_D^l}^{\top}=P\Sigma Q^{\top}
\end{equation}

\begin{equation}\label{eq_updD2}
\tilde{D}_{l}=QP^\top
\end{equation}

\begin{equation}\label{eq_updDh}
{h}_D^{l}=f_l(\tilde{D}_{l}{h}_W^{l})
\end{equation}
where the index $l$ increases from $1$ to $L-1$. As a classifier, the final output layer is updated by

\begin{equation} \label{eq_hWpd}
\tilde{W}_{L}={W}_{L}-\eta\frac{\partial \mathcal{L}({h}_W^L, \tilde{h}_W^L)}{\partial {W}_{L}}.
\end{equation}
Then the resulting $\{\tilde{W}_l, \tilde{D}_l\}_{l=1}^{L}$ constitute the final network. Note that the dictionary updating rules \eqref{eq_updD1} and \eqref{eq_updD2}  are adapted from  Algorithm \ref{alg-DicLearn}. For tractable computation, we simply  update  ${W}_{l}$ with the ridge regression model \eqref{eq_updW}, although other more sophisticated  models, such as the Lasso \cite{Tibshirani} and  elastic-net \cite{zou2005regularization}, probably yield better updates.

For large-scale data, such as ImageNet \cite{ImageNet09}, we suggest to process the data in batches and perform layerwise learning iteratively.     But for small or middle-scaled data, such as MNIST \cite{deng2012mnist} and CIFAR10 \cite{Krizhevsky09Cifar10}, the learning could be accomplished  in  a single pass. In this case,  compressed sensing-based target propagation needs to be performed only once, implying that the dictionaries are not required to be orthonormal in the final forward optimization and they could be  simply updated by
\begin{equation}\label{eq_updD}
\tilde{D}_{l}=\arg\min_{D_{l}} \|\tilde{h}_D^l-D_l{h}_W^{l}\|_2^2 +\lambda_{D_l}\|D_l\|_2^2,
\end{equation}
instead of  the SVD method used in \eqref{eq_updD1} and \eqref{eq_updD2}.





\section{Experiments}
In this section, we aim to prove that the proposed compressed sensing-based target propagation (CSTP) method could achieve better performance than the  auxiliary network-based target propagation (ANTP) method \cite{lee2015difference}, on the layerwise learning of deep networks. Moreover, we analyze  the proposed CSTP method by ablation study.  For  comparison, as in \cite{lee2015difference}, we test the proposed CSTP method on two typical datasets, MNIST \cite{deng2012mnist} and CIFAR10 \cite{Krizhevsky09Cifar10}. The MINIST dataset consists  of $28\times28$ gray-scale images of  handwritten digits.  It has a training set of 60,000 samples, and a test set of 10,000 samples. The CIFAR10 dataset consists of $32\times32$ natural color images in 10 classes. There are 50,000 samples in the training set and 10,000 samples in the testing set.


\subsection{Settings}
In the following, we briefly introduce the parameters set for the proposed CSTP-based layerwise learning algorithm. As shown in Algorithm  \ref{alg-layerTrain}, the algorithm   consists of three phases.   In the initial feedforward phase, the dictionaries are learned with Algorithm  \ref{alg-DicLearn}, for which we set the iteration number to be 20 and the sparsity parameter $d\approx m_l/3$. The thresholding function $f_l(\cdot)$ keeps about $k_l\approx m_l/2$ positive elements. By \eqref{eq_InitialWL}, we initialize the output layer $W_L$, namely the classifier. To obtain better performance, we suggest to handle  \eqref{eq_InitialWL} with stochastic gradient descent (SGD), for which we set the momentum  as 0.9,   weight decay as 0.0005,   batch size as 64, and  epoch number as 30. The learning rate is initialized as 0.01, which is decayed by multiplying a factor of 0.9 every 4 epochs. In the target backpropagation phase, we first  diversify the targets by  \eqref{eq_hDp}, with $\beta=0.2$.  Then  the targets are backpropagated by \eqref{eq_backD}, which is implemented with  OMP   \cite{Pati93}, with   $k_l\approx n_l/3$. In the final layerwise learning phase, we update  $W_l$  and $D_l$ by directly solving \eqref{eq_updW} and \eqref{eq_updD}.  The  output layer $W_L$ is updated in \eqref{eq_hWpd}  by SGD,  with the same parameters as set in \eqref{eq_InitialWL}. The above parameter settings are adopted both for MNIST and CIFAR10. 

\subsection{Results}

\subsubsection{Performance comparison on CIFAR10} For the layerwise learning with CSTP, we derive a test error of 47.76\% on a three-layer network of size $\text{3072-2500-1500-10}$. In contrast,  in \cite{lee2015difference}  ANTP reports an error  of 49.29\%  on a four-layer network of size $\text{3072-1000-1000-1000-10}$. This means that CSTP could achieve higher accuracy than ANTP, when given appropriate network sizes.  Recall that the network size of CSTP is constrained by compressed sensing. Similarly to  ANTP \cite{lee2015difference}, the CSTP-based layerwise learning  performs  worse than the prevailing BP method, with an accuracy gap about 4\%  on the network we test.  As discussed in \cite{ma2020layer}, the inferior performance is due to the mismatching between the poorly-separable features  and the strong supervision constraint imposed on each layer.

\subsubsection{Performance comparison on MNIST} In \cite{lee2015difference},  ANTP achieves a test error of 1.94\% on a network with 7 hidden layers each consisting of 240 units. The error is reduced about  0.15\%, as we test CSTP on a wide but shallow network of size  $\text{784-3500-2500-10}$. From Table 1, it is seen that  wider  layers tend to yield better  performance for CSTP. The reason is as follows. The performance of CSTP depends on the dictionary learning-based feature selection and   compressed sensing-based target propagation. For dictionary learning, higher dimensions (namely wider layers) usually could lead to sparser transforms, which are beneficial not only to feature selection but also to compressed sensing.

\subsubsection{Ablation study of $D_l$ and $f_l(\cdot)$}
To see the importance of the dictionary $D_l$ and nonlinear function $f_l(\cdot)$ in CSTP,  we remove them in Algorithm \ref{alg-layerTrain}, and after fine tuning, we still witness an accuracy reduction about 1.2\% for CIFAR10 and   0.7\% for MNIST. If only removing $f_l(\cdot)$, the accuracy decreases about 0.3\% for CIFAR10 and 0.1\% for MNIST.

   \begin{table}[t]
   \caption{The layerwise learning performance  of  CSTP on different network architectures. \label{tab:NetSize}}
		\centering
        \footnotesize{
		\resizebox{0.8\columnwidth}{!}{
		\begin{tabular}{c|c|c}\hline
		  Datasets & Network sizes &  Error ($\%$)\\ \hline
       \multirow{3}{*}{MNIST}&784-4000-3000-10& 1.72\\
                        &784-3500-2500-10& 1.79\\
                        &784-2000-1000-10& 2.86\\\hline
       \multirow{3}{*}{CIAFR10}&3072-4000-3000-10&47.48\\
                        &3072-2500-1500-10& 47.76\\   &3072-2000-1000-10& 48.87\\\hline
		\end{tabular}
		}
}
	\end{table}

\section{Conclusion}
In the letter, we have shown that the cascade of compressed sensing can be established as a reversible network, which allows  for layerwise learning and analysis. The optimization target of each layer is derived by inversely propagating the target of the network via compressed sensing. The method avoids learning an auxiliary network to specially propagate the target,  reducing the complexity of network learning. Also, the method could achieve favorable performance owing to the stableness of compressed sensing.  The proposed target propagation involves two major computations: dictionary learning and sparse recovery, which both have polynomial complexity. To reduce the complexity, it is interesting to sparsify  the structures of the dictionaries and projection matrices, such that some combinatorial algorithms with linear complexity could be adopted \cite{Gilbert10,lu2018expander}. This will be left for our future work. Note that in the letter we focus our interest on the layerwise learning method that optimizes exactly one layer at a time and thus reduces the network analysis to the layer level. Currently, the method seems difficult to achieve BP level performance, especially on large-scale data \cite{ma2020layer}. The reason is that the finite number of parameters in one layer can hardly handle  the matching between poorly-separable features and strong supervision constraints \cite{ma2020layer}. The problem could be addressed if more than one layer is learned at a time \cite{belilovsky2019greedy,nokland2019training}.

\bibliographystyle{IEEEtran}
\balance
\bibliography{refs_20210701}
\end{document}